%% file: acl_latex.tex
\pdfoutput=1

\documentclass[11pt]{article}

\usepackage[]{acl}

\usepackage{times}
\usepackage{latexsym}
\usepackage{graphicx}
\graphicspath{ {./figures/} }
\newcommand*{\affaddr}[1]{#1} 
\newcommand*{\affmark}[1][*]{\textsuperscript{#1}}

\usepackage{xcolor}
\usepackage{times,latexsym}
\usepackage{url}
\usepackage[T1]{fontenc}
\usepackage[utf8]{inputenc}

\usepackage{ocr}
\usepackage{microtype}

\usepackage[smallerops,vvarbb]{newtxmath}

\usepackage{xfrac}

\usepackage{algorithm}
\usepackage[noend]{algpseudocode}

\algnewcommand{\LineComment}[1]{\State \(\triangleright\) #1}

\algblockdefx[CommentHead]{CommentHead}{EndCommentHead}[1]{\textcolor{gray}{\(\triangleright\) #1}}{}
\algtext*{EndCommentHead}

\algblockdefx[Class]{Class}{EndClass}[1]{\textbf{class} #1}{}
\algtext*{EndClass}

\algnewcommand\algorithmicbreak{\textbf{break}}
\algnewcommand\Break{\State \algorithmicbreak}
\algnewcommand\algorithmiccontinue{\textbf{continue}}
\algnewcommand\Continue{\State \algorithmiccontinue}

\algnewcommand\algorithmicinput{\textbf{Input:}}
\algnewcommand\algorithmicoutput{\textbf{Output:}}
\algnewcommand\Input{\item[\algorithmicinput]}
\algnewcommand\Output{\item[\algorithmicoutput]}

\DeclareMathAlphabet{\mathcal}{OMS}{cmsy}{m}{n}

\definecolor{sjmlime}{rgb}{0.9,1,0.3}
\definecolor{bsgrey}{rgb}{0.9,0.9,0.9}
\usepackage{xspace}
\usepackage{todonotes}
\makeatletter
\newcommand*\iftodonotes{\if@todonotes@disabled\expandafter\@secondoftwo\else\expandafter\@firstoftwo\fi}  
\makeatother

\newlength{\extramargin}
\usepackage{calc}

\usepackage{url}
\usepackage{colortbl}

\usepackage{tikz}
\usetikzlibrary{arrows,matrix,positioning,fit,calc,shapes,decorations.pathreplacing,svg.path,patterns}
\pgfdeclarelayer{background}
\pgfdeclarelayer{backgroundoverlay}
\pgfsetlayers{background,backgroundoverlay,main}
\usepackage{pgfplots}
\pgfplotsset{compat=1.13}  

\usepackage{nicefrac}
\usepackage{enumerate}

\usepackage{adjustbox}

\usepackage[capitalise]{cleveref}

\crefname{section}{\S}{\S\S}
\Crefname{section}{\S}{\S\S}
\crefformat{section}{\S#2#1#3}  

\usepackage[T1]{fontenc}

\usepackage[utf8]{inputenc}

\usepackage{microtype}

\title{\Large{Masader: Metadata Sourcing for Arabic Text and Speech Data Resources}}

\author{
Zaid Alyafeai\affmark[1], Maraim Masoud\affmark[2], Mustafa Ghaleb\affmark[1], and Maged S. Al-shaibani\affmark[1]\\ \\
\affaddr{\affmark[1] King Fahd University of Petroleum and Minerals, Dhahran, Saudi Arabia}\\
\affaddr{\affmark[2] Independent Researcher}\\
}

\begin{document}
\maketitle

\begin{abstract}
The NLP pipeline has evolved dramatically in the last few years. The first step in the pipeline is to find suitable annotated datasets to evaluate the tasks we are trying to solve.  Unfortunately, most of the published datasets lack metadata annotations that describe their attributes. Not to mention, the absence of a public  catalogue that indexes all the publicly available datasets related to specific regions or languages. When we consider low-resource dialectical languages,  for example, this issue becomes more prominent. In this paper we create \textit{Masader}, the largest public catalogue for Arabic NLP datasets, which consists of 200 datasets annotated with 25 attributes. Furthermore, We develop a metadata annotation strategy that could be extended to other languages. We also make remarks and highlight some issues about the current status of Arabic NLP datasets and suggest recommendations to address them.
\end{abstract}

\input{1_introduction}
\input{2_related_work}
\input{3_methodology}
\input{4_data_and_search_results}
\input{5_examining_arabic_nlp_landscape}

\input{51_issue_and_recommendation}

\input{6_limitations_and_future_work}
\input{7_conclusion}
\input{8_acknowledgements}

\bibliography{anthology,custom}
\bibliographystyle{acl_natbib}

\appendix
\section{Extra Analysis}

\begin{figure}[ht!]
\includegraphics[width=0.45\textwidth]{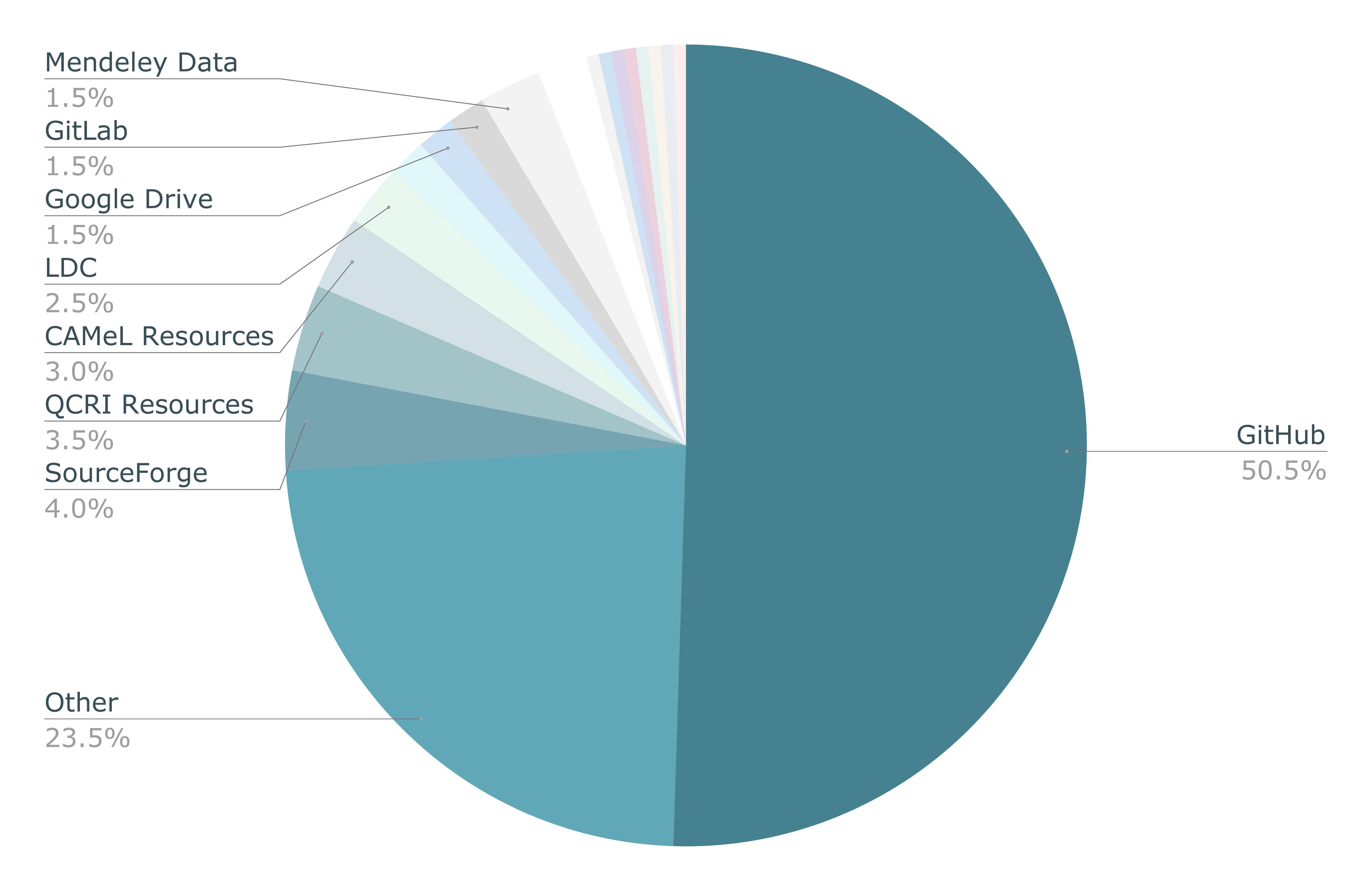}
\caption{Most used repositories to host datasets.}
\label{fig:repos}
\centering
\end{figure}

\paragraph{Repositories} In Figure \ref{fig:repos}, we highlight the most used repositories to host datasets. More than 50 \% of the datasets are hosted on GitHub. While around 23 \% are hosted on arbitrary websites. We notice that two main university resources are used which are QCRI (Qatar Computing Research Institute) and CAMeL (Computational Approaches to Modeling Language Lab). On the other hand most of the paid resources are on LDC (Linguistic Data Consortium). There are also other free websites for data hosting including SourceForge, GitLab and Mendeley Data. A few percentage of the datasets are also hosted in Google Drive and Dropbox. 

\begin{figure*}[htp!]
\includegraphics[width=1\textwidth]{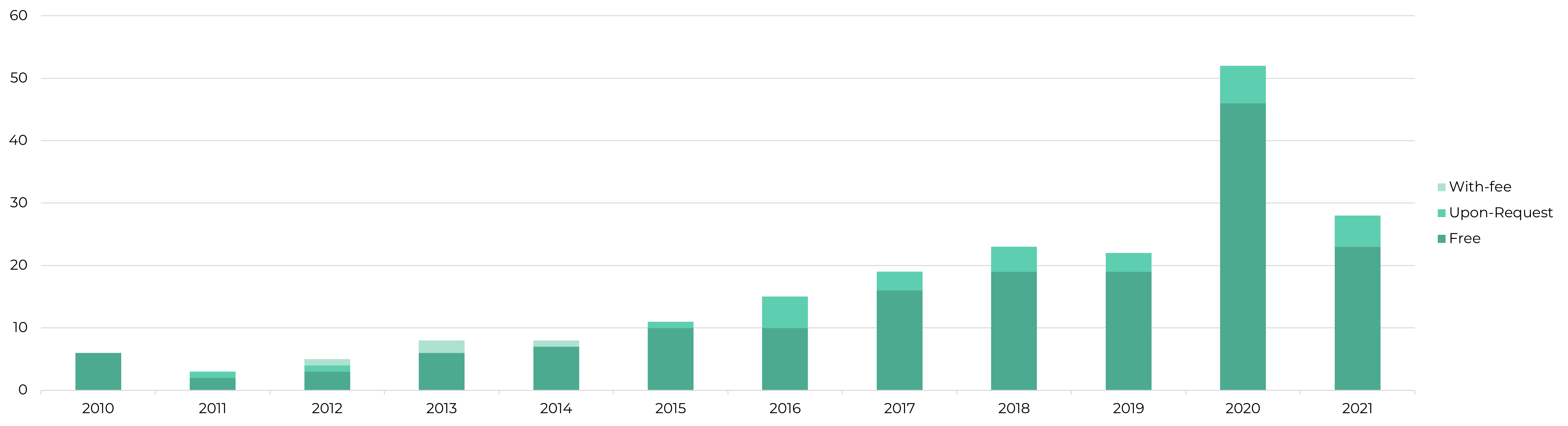}
\caption{Breakdown of the cost associated with the datasets from 2010 to 2021.}
\label{fig:year_access}
\centering
\end{figure*}

\begin{figure*}[htp!]
\includegraphics[width=1\textwidth]{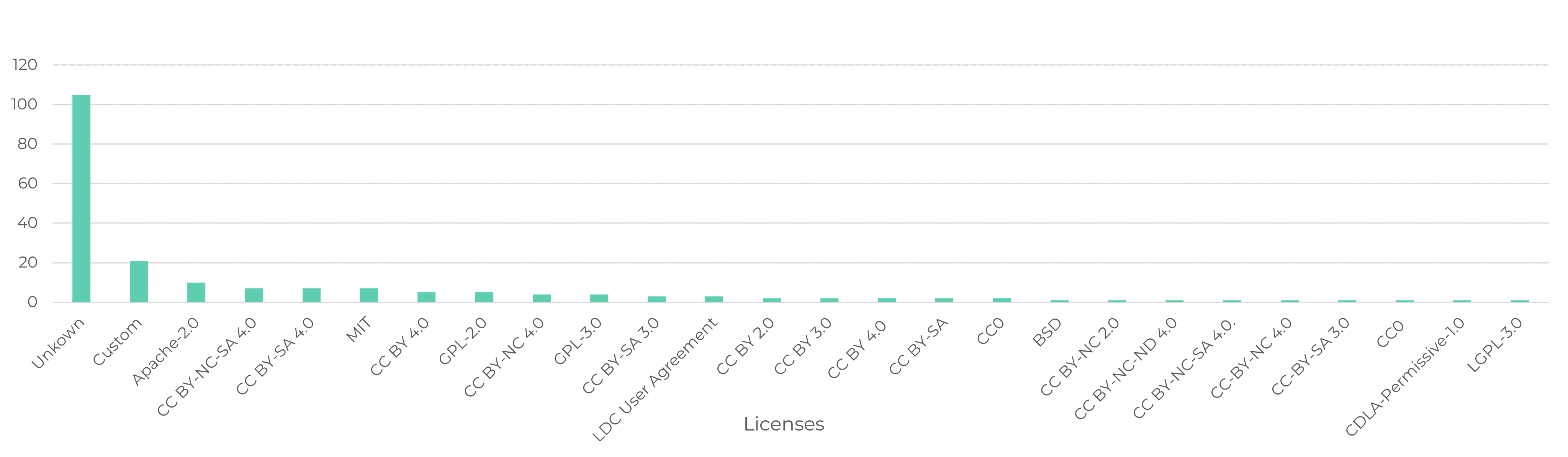}
\caption{Distribution of licenses across the datasets.}
\label{fig:license}
\centering
\end{figure*}

\paragraph{Accessibility} In Figure \ref{fig:year_access}, we breakdown the three types of accessibility for datasets. We highlight that most of the datasets are free with a small percentage that either requires registration or a fee (mostly LDC). We observe a general trend of mainly publishing free data (around more than 80 \% in the last few years).  In the repositories that host data, we observe that more than 50 \% of the data providers don't declare the type of the license for the datasets as stated in Figure \ref{fig:license}. On the other hand around 10 \% use custom licenses. Typically the most used standard licenses are variations of Creative Common, followed by Apaache, GPL and MIT.  

\begin{figure*}[htp!]
\includegraphics[width=1\textwidth]{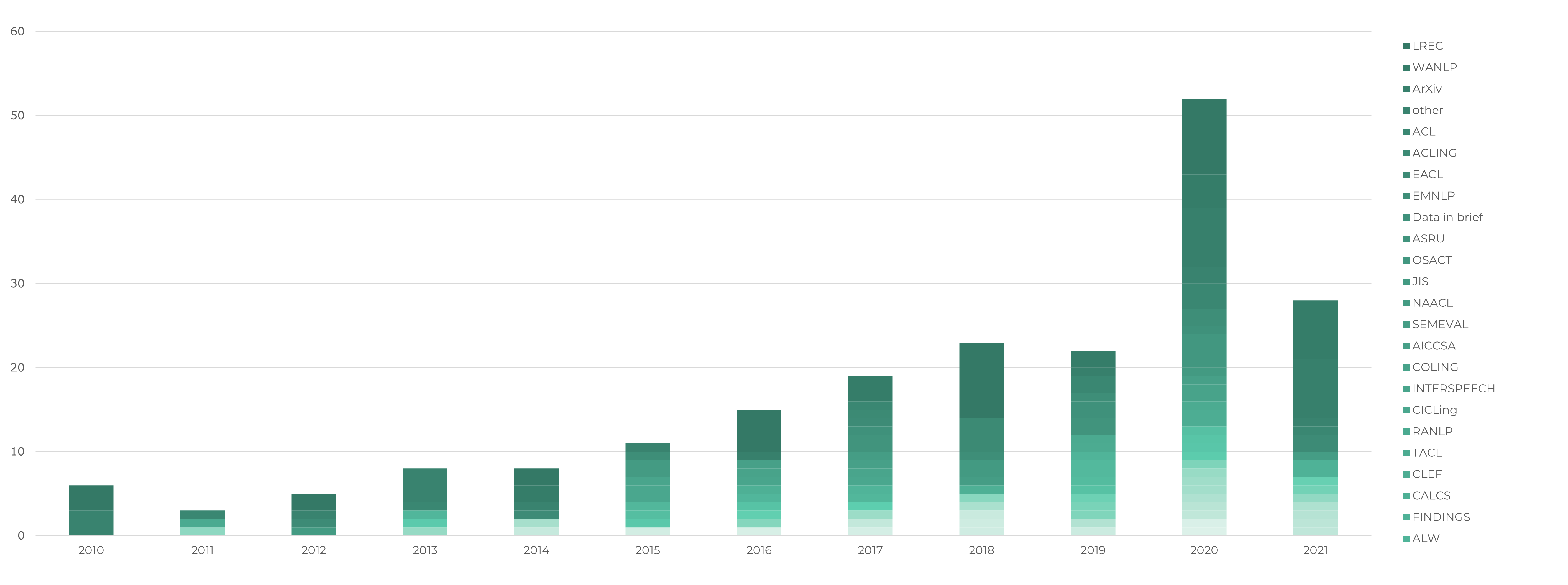}
\caption{The count of venues within each year. }
\label{fig:year_venue_bd}
\centering
\end{figure*}

\paragraph{Venues} Figure \ref{fig:year_venue_bd} breakdowns the venues that are used to publish the datasets across the different years. We observe  variations in the venues with around 70 unique venues across conferences, journals, workshops and preprints. As we observe from the figure the most used ones are LREC, WANLP followed by preprints (including arXiv and others). The top venues are mainly conferences and workshops.

\begin{figure*}[htp!]
\includegraphics[width=1\textwidth]{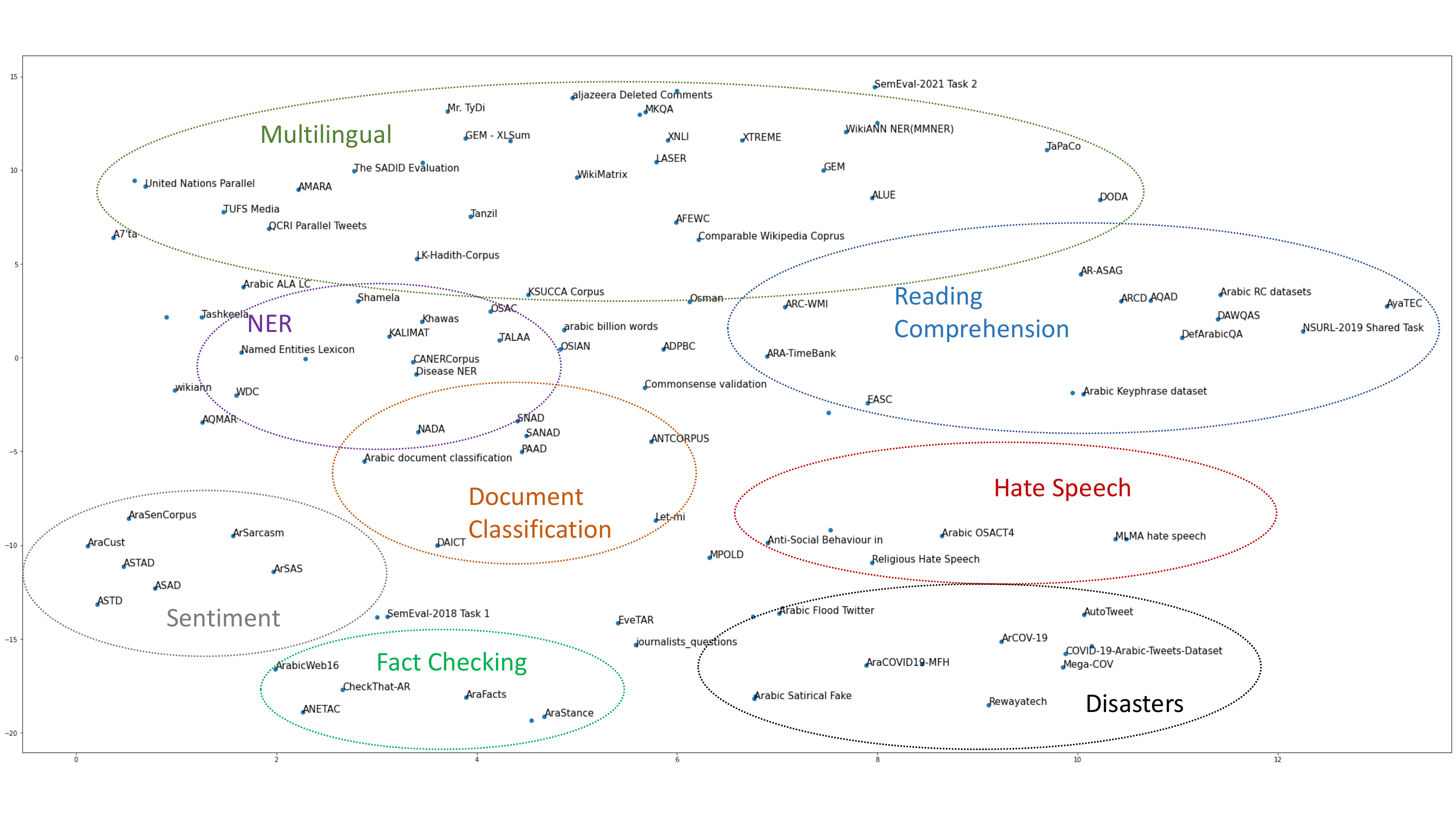}
\caption{Positive projected embeddings of all datasets' abstracts. }
\label{fig:embeddings_pos}
\centering
\end{figure*}

\begin{figure*}[htp!]
\includegraphics[width=1\textwidth]{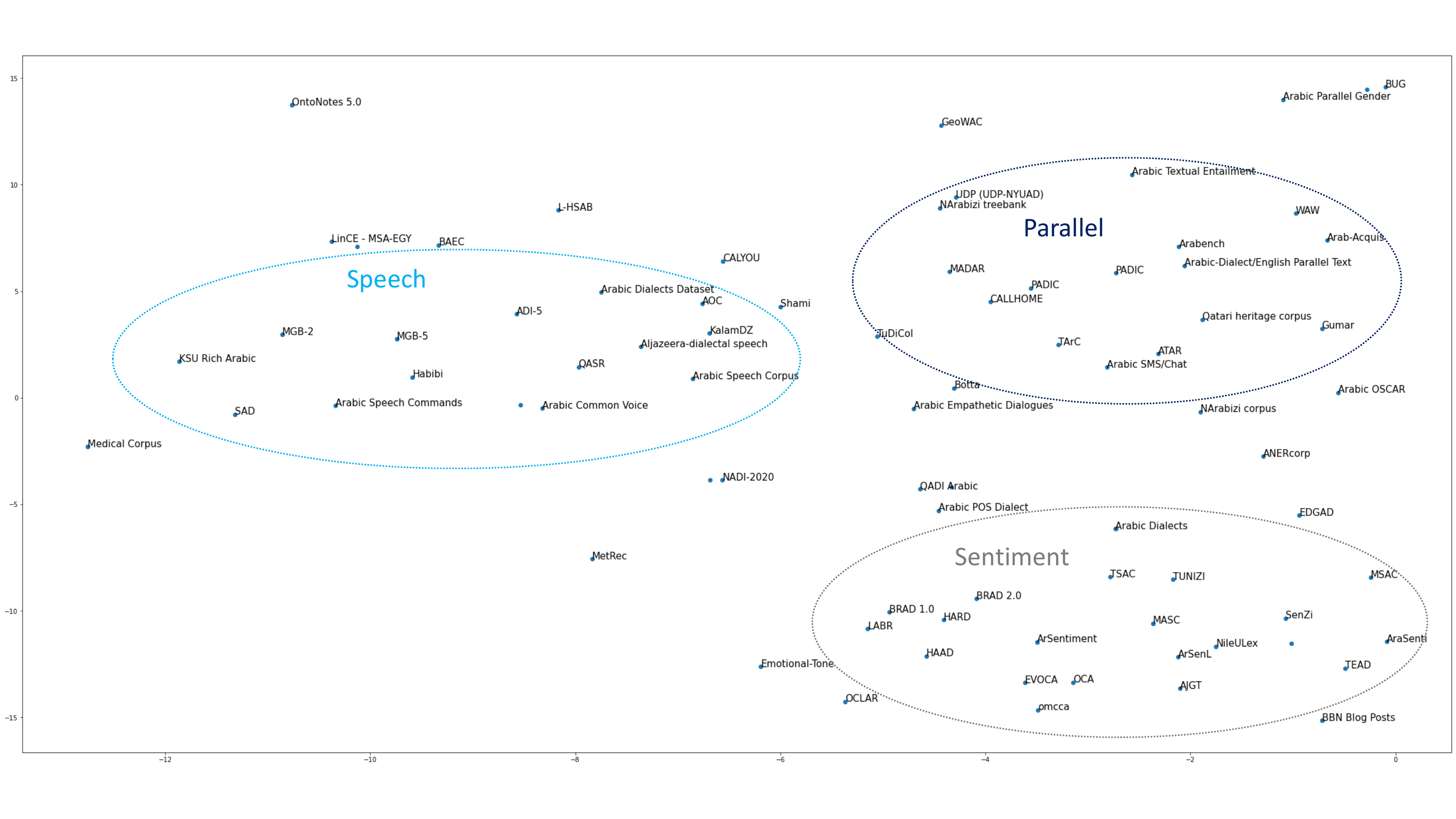}
\caption{Negative projected embeddings of all datasets' abstracts. }
\label{fig:embeddings_neg}
\centering
\end{figure*}

\paragraph{Abstract Projection} In Figures \ref{fig:embeddings_neg} and  \ref{fig:embeddings_pos}, we highlight all the datasets that we collected as projected embeddings. The embeddings were created by extracting the abstracts of all the datasets using Semantic Scholar API then project the sentence embeddings extracted from sentence-transformers \footnote{\url{https://github.com/UKPLab/sentence-transformers}}. The embeddings are of shape (200, 384), were projected to the 2D space using the t-SNE algorithm \cite{van2008visualizing}. We separate the positive and negative embeddings of the x-axis for better visualization and analysis. We manually highlight 11 clusters for datasets that share common attributes. The clusters could be grouped by task (like sentiment analysis), format (like speech) or type (like parallel or multilingual datasets). 

\begin{figure}[htp!]
\includegraphics[width=0.48\textwidth]{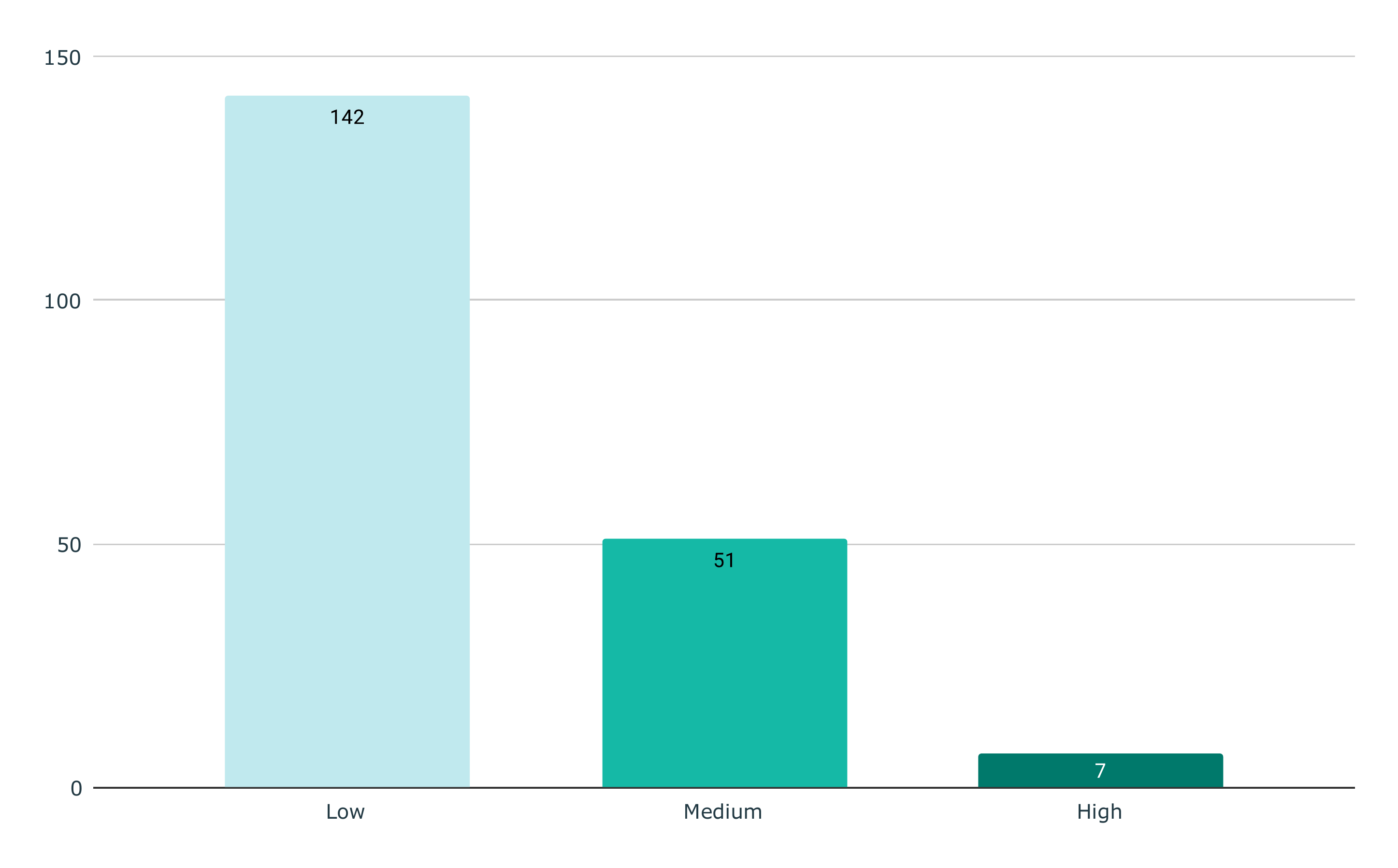}
\caption{Distribution of datasets in terms of ethical risks.}
\label{fig:ethical}
\centering
\end{figure}

\paragraph{Ethical Risks} Figure \ref{fig:ethical} highlights the distribution of datasets in terms of ethical risks. Datasets that could potentially contain personal information are labeled as medium. On the other hand, datasets that might contain, additionally, toxic information or hate speech text are considered as high ethical risks. All the remaining datasets are considered low. As we can observe from the Figure, most of the collected resources have low ethical risks with around 30 \% having medium or high ethical risks. 
\paragraph{Domain Representation}
In language modelling, it is important to include datasets that are representative not just of language diversity, but also of the domains or topics covered by the datasets. In Table \ref{tab:domain}, we highlight the various domains presented in the surveyed datasets. The majority of the datasets cover a variety of domains, because they could be scrapped from the web, Wikipedia, or collected manually.  Around 30 \% of the datasets are from social media, with the remaining 12 \% coming from news articles. Books and reviews account for just a minor fraction of the genres in the datasets.

\begin{table*}[htp!]
\caption{Summary of domains in the surveyed datasets. }
\label{tab:domain}
\begin{tabular}{l|c}
\hline
\textbf{Domain} & \textbf{Count}  \\ \hline \hline
social media &  61 \\ \hline
news articles & 24 \\ \hline
transcribed audio & 16 \\ \hline 
reviews & 9  \\ \hline 
books & 7\\ \hline  \hline 
other & 83 \\ \hline 
\end{tabular}
\centering
\end{table*}

\end{document}

%% file: 1_introduction.tex
\section{Introduction}



The emergence of deep learning and its applications in many fields had a great impact on the development of various natural language processing (NLP) and speech techniques that were adapted to many languages. Many might correlate that to the availability of data especially with the existence of social media and the manufacturing of hardware devices that fostered research in the field, namely GPUs. Typically, we are referring to  the era of deep learning which started roughly after 2010. Following that, many public Arabic NLP and speech datasets have been published in conjunction with the recent advances in deep learning \cite{DBLP:Zaghouani17}. Currently, there is no online centralized catalogue for Arabic NLP and speech  datasets. It is unclear how many online datasets there are as well as the metadata describing the datasets' characteristics, such as diversity, demographic distribution, ethical considerations, quality, and so on. This study attempts to identify the publicly available Arabic NLP datasets and to provide  a catalogue of  Arabic datasets to researchers. The catalogue will increase the discoverability and provide some key metadata that will help researchers identify the most suitable dataset for their research questions. 

We highlight our contributions as the following: 
\begin{itemize}
    \item We create the largest catalogue with 25 attributes for 200 Arabic NLP and speech datasets. 
    \item We design a metadata schema for annotating the datasets. 
    \item We analyse the current status of the Arabic NLP and speech datasets, discover issues and recommend solutions. 
    
\end{itemize}
The paper is structured as follows. Section \ref{related_work} looks into previous work in the literature.  Section \ref{methodology} summarizes our approach to develop the catalogue. It discusses  the research methodology, metadata design, and the annotation process.  Section \ref{findings} outlines our findings. The results are then inspected, and issues and recommendations are highlighted in sections \ref{rq1_rq2} and \ref{rq3} respectively.

%% file: 2_related_work.tex
\section{Related Work}
\label{related_work}


Surveying the literature to derive analysis about a specific research field or topic is a standard practice. It helps to provide an overview on the directions and trends on subject of interest. A prominent example of such effort is \citet{ammar2018construction}. They collected a large corpus of 280 million nodes. These nodes are diverse entities representing authors, papers, etc. An application of their work is the connected papers project \citet{connected_papers}. It aims to construct a graph of related literature based on a given query. \citet{radev2016bibliometric} extends the analysis by accounting for the citation count and extensive manual annotation on the collected literature. They curated their dataset from ACL Anthology papers. Their analysis covers various various attributes including authors impact factor, h-index and collaboration. For massive analysis reports on the field of NLP, \citet{mohammad-2020-nlp} surveyed the literature with 1.1 million paper information dataset collected from Google Scholar. Additionally, \citet{sharma2021drift} proposed DRIFT, a data analysis tool that presents an overview of the landscape of a queried topic. They constructed their dataset from arXiv papers' abstracts. 

As the dataset collection research is vastly growing with the significant emergence of data on the web, the need to mandate such process becomes a necessity. This is, in fact, an active branch of research happening across various domains and disciplines. Consider, for instance, the systematic literature review protocol developed by \citet{kitchenham2004procedures} that governs the data collection process for surveys. Another example is the guidelines reported by \citet{mbuagbaw2017considerations} on clinical trials. There are also studies that propose standardizing the documentation of datasets. \citet{gebru2018datasheets} proposed datasheets for datasets. Their aim is to accompany datasets with a descriptive datasheet schema describing diverse attributes about the dataset. Such attributes include operating characteristics, recommended uses, motivation, collection process and test results. Similarly,  \citet{bender-friedman-2018-data} propose data statement, a similar standardization approach that overlaps to datasheets in some of its attributes. However, while datasheets aims to document more general information about the dataset, data statement is more specific to linguistics and NLP.

In the field of Arabic datasets, there are many studies that attempted to survey the available data resources. \citet{shoufan-alameri-2015-natural} reviewed the NLP literature for dialectical Arabic. 
Their work can be considered as a quick reference to locate important contributions for certain Arabic dialects that address specific NLP features. However, this study reviewed limited literature as it was concentrated only on Arabic dialects and the research on Arabic corpora was still infant. A comprehensive approach was implemented by \citet{DBLP:Zaghouani17} where they collected a list of around 80 freely available Arabic datasets including datasets that are not related to NLP. The also provide links to the datasets but some of them do not work anymore. There are also some efforts to survey specific dialects. For example, \citet{younes2020language} provided a review of various kinds of constructed language resources (LRs) of Maghrebi Arabic dialects (MADs). They reviewed MAD raw corpora and divided it into speech corpora, speech transaction, web, and social media corpora. Recently, \citet{guellil2021arabic} presented and classified 90 studies that covered classical Arabic, Modern Standard Arabic, and Arabic Dialects. Also, they provided links to around 52 NLP datasets. Another survey paper was published by  \citet{Darwish2021APS} to review the available tools and resources for Arabic. However, they provided links to a few datasets.  




%% file: 3_methodology.tex
\begin{figure*}[t]
\includegraphics[width=\textwidth]{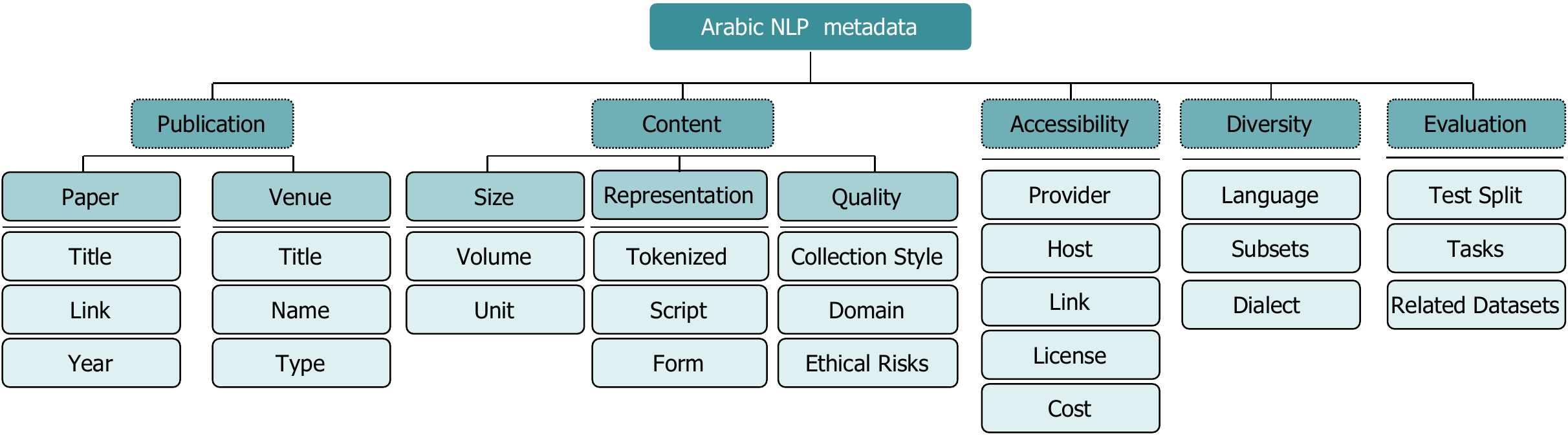}
\caption{Metadata schema for Arabic NLP resources.}
\label{fig:metadata}
\centering
\end{figure*}

\section{Methodology}
\label{methodology}
The research method behind this survey follows the keyword-base literature review process \citet{rowley2004conducting} followed by an annotation process to enrich the filtered results with metadata.
Our methodology follows five steps: (i) searching resources, (ii) filtering using a selection criteria, (iii) annotating resources with metadata, (iv) validating the resources, and (v) analysing the results.


Based on a pre-selected keywords, the retrieved Arabic language resources are added to our \textit{preliminary list} of data sources. Next, all the resources collected are filtered according to a set of inclusion criteria. The datasets that passed the filtering criteria are ported to our \textit{final list} of datasets, and then are annotated with a set of metadata, both manually and automatically. Those which do not pass the criteria are discarded. Following that, a verification step is performed to ensure the accuracy of the metadata. Finally, the final set of resources are analysed according to the metadata and presented in this study. 

The initial search and the filtering for all sources were done between July and August 2021, and the annotation process took place progressively after the data was collected and concluded on September 2021. The following subsections describe each step in more detail.

\subsection{Step 1: Searching Resources}
The targeted search is performed using Google Search Engine to identity Arabic NLP dataset directly. To search specific dataset, we also conduct the search  against the well-known data repositories and indexing websites using a set of keywords. The selected repositories are GitHub\footnote{\url{https://github.com/}}, Paperswithcode\footnote{\url{https://paperswithcode.com/}}, Huggingface\footnote{\url{https://huggingface.co/}}, LREC\footnote{The International Conference on Language Resources and Evaluation \url{http://www.lrec-conf.org/}}, Google Scholar\footnote{\url{https://scholar.google.com/}}, and  LDC\footnote{The Linguistic Data Consortium \url{https://www.ldc.upenn.edu/}}. The search combined terms related to NLP and Arabic language, such as "\textit{NLP}", "\textit{Natural Language Processing}" and all variations of Arabic dialects, as well as terms such as "\textit{database}", "\textit{dataset}", "\textit{resource}", and  "\textit{corpus}".
This step generates our \textit{preliminary list} of data sources which consists of around 299 resources.

\subsection{Step 2: Filtering with Inclusion Criteria}
Our search was additionally supplemented by manually screening retrieved articles and datasets, which we perform using  a set of inclusion criteria, which are as follows:
\begin{itemize}
    \itemsep0em 
    \item the dataset is specific for NLP.
    \item there is a publication associated with the language resource. 
    \item the resource is created after 2010.
    \item the resource, in its raw form or annotated format, is suitable for language modelling and text generation tasks. 
    \item the resource does not serve as NLP tool such as spelling-checking and stop-words.
\end{itemize}

From the \textit{preliminary list}, we ended up with 207 papers. Our inclusion criteria removed 93 papers initially. 

\subsection{Step 3: The Annotation Process}
By applying the inclusion criteria, the datasets search pool is reduced to the final set of considered resources. At this stage, an annotation process is applied to manually annotate them with a set of pre-agreed metadata.  When applied to this process, we consider three main goals: 1) designing metadata specific for Arabic resources, 3) deciding on the annotation format, 2) setting up an annotation workflow, and finally 3) defining an annotation task.

\paragraph{Metadata Selection}
The main motivation behind designing metadata for Arabic NLP resources is to increase the  discoverability and reusability of such resources. The metadata are chosen to represent different aspects of the language resource.
\citet{Park2021KLUEKL}'s work serves as an example for identifying the appropriate metadata for our Arabic NLP use case. Following several revisions, the final agreed-upon metadata is represented by a taxonomy in Figure \ref{fig:metadata}. It consists of five subcategories as follows:

\begin{itemize}
    \itemsep0em 
    \item \textit{Publication}: This subcategory concerns about metadata relating to author, publisher and other publication details for the dataset referenced publication. It includes attributes such as the \textit{title}, the \textit{link}, the \textit{year} of publication for the referenced paper, and the venue \textit{title}, \textit{name}, and \textit{type}.
    
    \item \textit{Content}: This subcategory is concerned  with the content of the dataset in terms of the size, the representation and the quality. The size tag indicates the quantity of the dataset by specifying  the \textit{unit} of measurement (tokens, sentences, documents, MB, GB, TB, hours, others), as well as  the number of units using the \textit{volume} tag. The representation dimension describes the contextual information about the dataset. For example, the \textit{tokenized} flag specifies whether the dataset is tokenized. This is useful since various tokenizers project different behaviour, and when this is not specified, it impacts downstream tasks. The \textit{form} tag, on the other hand, defines the form of the content, being written, or spoken  language, while the \textit{script} tag describes the writing system used in the dataset (Arab,Latn,Arab-Latn,Other). The third dimension, quality, describes elements related to the data collection.  It covers, for example,  the \textit{collection style} used for building the dataset (e.g crawling, translation, etc), the \textit{ethical risk} associated with utilizing the data set (low, medium, and high), and  the \textit{domain} of the dataset (social media, etc ).

    \item \textit{Accessibility}: This subcategory concerns about the timeliness and the reliability of access to the data. Its associated metadata includes: the name of the data \emph{provider}, the name of the data \emph{host}, the \emph{link} to download the data from the host, the \emph{licence} and the \emph{cost} to obtain the data. 
    
    \item \textit{Diversity}: This metadata subclass is used to capture the linguistic and culture diversity within Arabic language. It covers the \emph{language} tag to represent the language of the dataset, either being Arabic (\emph{ar} or \emph{multilingual} to denote a dataset that contains several languages), as well as the \textit{subsets} tag to denote the sub-datasets that are contained inside this dataset. The last tag in this class is \textit{dialect}. To capture the linguistic variety of Arabic, we adapted five high-level categories of dialect variations, resulting in a total of 29 dialect categories. These categories are as follows: i) MSA for Modern Standard Arabic, ii) CLS for Classic Arabic and Qura'anic text, ii) Regional dialects for the four regions (GLF, LEV, EGY, NOR)\footnote{GLF (Gulf region), LEV (Levant region), EGY (Egypt and Sudan) and  NOR (North Africa region).}, iii) country-based dialects which cover the 22 dialects spoken in Arabic-speaking countries, and finally iv) the other which includes mixed dialects and code-switched script.
     
    \item \textit{Evaluation}: The metadata within this subcategory describes the process of using the dataset in relation to the evaluation phase of the NLP pipeline.The first tag is the  \textit{test split}, which is deployed as a boolean flag to signal if  the dataset is prepared for evaluation task by having a distinct split between the training and the test sets. The \textit{tasks} tag defines the list of tasks to which the dataset is applied, whilst the \textit{related datasets} attribute, lists what dataset(s) intersect with the current dataset.
\end{itemize}

\paragraph{Annotation Formats}
Based on the chosen metadata, we figured out that different annotation procedures can be applied to insert the metadata. Hence, two formats of annotation are adapted in this work; (i) manual curation, and (ii) automatic annotation of the metadata. 
The manual curation is performed by the human annotators via manually inspecting the dataset link, and it's referenced paper. We basically use this format to extract metadata that is hard to automate or not implicitly mentioned. The second format is auto-annotation. For this format, we rely on APIs from academic publishers, such as Semantic Scholar API (Python Library) \footnote{\url{https://api.semanticscholar.org/graph/v1}}. As such, most of the metadata is manually annotated, except for the publication information which was retrieved using the API. 


\paragraph{Annotation Workflow} We used Google Sheets to set up our annotation workflow, where the metadata is specified as Google Sheet columns. The datasets were annotated with a set of metadata by the four authors, who are fluent Arabic speakers and researchers in the field of natural language processing. 

We define the manual annotation task as follows. For each link of Arabic language resource in our filtered pool of resources, the main goal of the task is to annotate the filtered datasets against the metadata. The annotators were instructed to follow the following set of guidelines:
\begin{enumerate}
\itemsep0em 
    \item Examine the resource link.
    \item Examine the referenced paper.
    \item Fill the metadata entries on Google Sheets.
    \item If a particular attribute of the dataset is not mentioned, log it in the notes' column.
    \item If a conflict is observed between the reported metadata from the resource link and the actual published paper, mark this entry in the sheet.
\end{enumerate}

\subsection{Step 4: Verification of metadata}
Following the completion of the annotation, a verification step is performed to confirm the accuracy of the information given. This step was deemed necessary in order to explain the notes from the prior stage. It is done manually and involved active communication amongst the annotators. At this stage, we removed 7 extra papers, 2 of them were duplicated datasets and 5 had wrong annotations for the year attribute (before 2010). In Figure \ref{fig:example}, we show an example of the metadata annotations of a chosen dataset. 

\subsection{Step 5: Analysis }
After verifying the final collection of annotated datasets, we conduct the analysis on various metadata.  
The findings of the analysis will be described in the next sections.

%% file: 4_data_and_search_results.tex
\begin{figure*}[t]
\includegraphics[width=\textwidth]{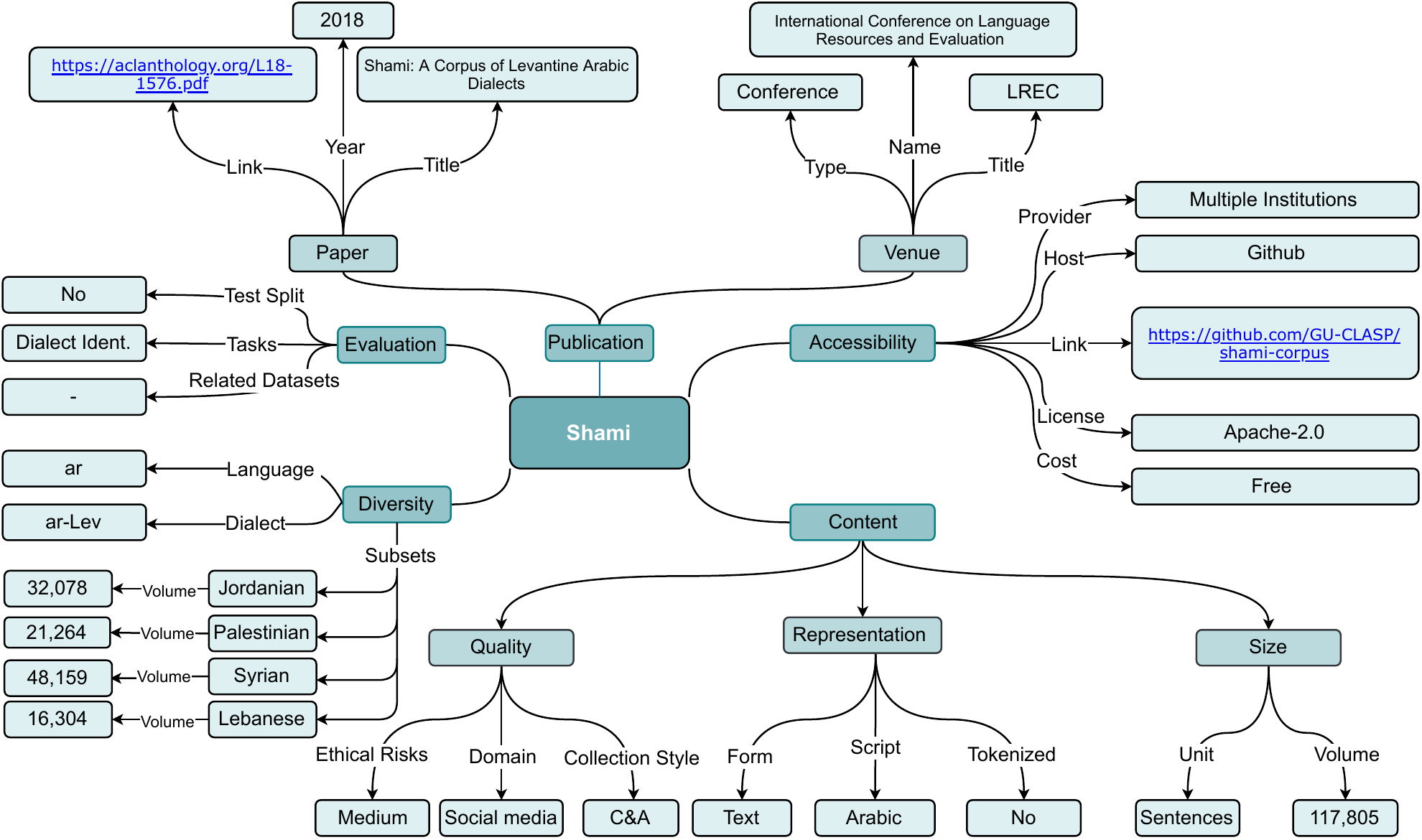}
\caption{Example demonstrates the annotation of the metadata on the Shami dataset \cite{abu-kwaik-etal-2018-shami}. The \textit{subsets} tag represents the dialects and each subset (For example, Jordanian) inherits all the metadata from the superset Shami, except the \textit{volume}.}
\label{fig:example}
\centering
\end{figure*}

\section{Findings and Representation}
\label{findings}

In this section we describe our findings in collecting all the data resources related to Arabic NLP published between 2010 and September 2021. We describe driven statistics of the datasets in addition to how we represent our data in a user friendly format. 

\subsection{Data Statistics}
The total number of datasets included in this catalogue is 200. More than 90 \% of the datasets' written format is text while the remaining is speech data. Table \ref{tab:volume} summarizes the overall statistics of the catalogue in terms of volume. We mainly used the reported numbers in the paper and validated the numbers by downloading the dataset. If the size is different we report the numbers from the downloaded dataset. If the number can't be validated, because of the size of the dataset for example, we report the numbers from the paper. Mostly, we use tokens to represent datasets that tackle token-based tasks like named entity recognition (NER), sentences to represent datasets that are related to sentence-based tasks like sentiment analysis and documents if the size of the dataset is too large. 

\begin{table}[!htp]
\caption{Summary of the 200 Arabic NLP datasets in the Masader project in terms of volume.}
\label{tab:volume}
\begin{tabular}{l|c}
\hline
\textbf{Unit} & \textbf{Volume}  \\ \hline \hline
Tokens &  451,370,314 \\ \hline
Sentences & 1,236,350 \\ \hline
Documents & 51,701 \\ \hline 
Hours & 3,104.1  \\ \hline  \hline
\# Datasets & 200 \\ \hline 
\# Datasets with Dialect Subsets & 23 \\ \hline 
\# Total Subsets & 375 \\ \hline 
\end{tabular}
\centering
\end{table}

\subsection{Masader Interface}
To easily navigate the sources we created a website \footnote{\url{https://arbml.github.io/masader/}} that is connected directly to the Google Sheets, allowing any updates in the sheets to be reflected immediately on the website. The website's primary interface only displays  nine attributes\footnote{index, name, link, year, volume, unit, paper link, access, and tasks.}. These are deliberately chosen for trail, and testing their relevance for academic search. The interface supports discoverability by including the following features: 1)  a clickable association between the dataset and its published paper, 2) a direct link to the most recent hosted version of the dataset, 3) a clickable link on the dataset name, which leads to dataset card displaying the remaining metadata of a dataset,  and finally 4) filtering and sorting based on each attribute.


%% file: 5_examining_arabic_nlp_landscape.tex
\section{Examining the Arabic NLP Landscape}
\label{rq1_rq2}
This section provides an analysis on the surveyed datasets. We mainly focus on discussing the current trend of publishing Arabic resources and drawing some remarks about the overall status of the landscape. 

\subsection{Publications development} 

\begin{figure*}[t]
\includegraphics[width=1\textwidth]{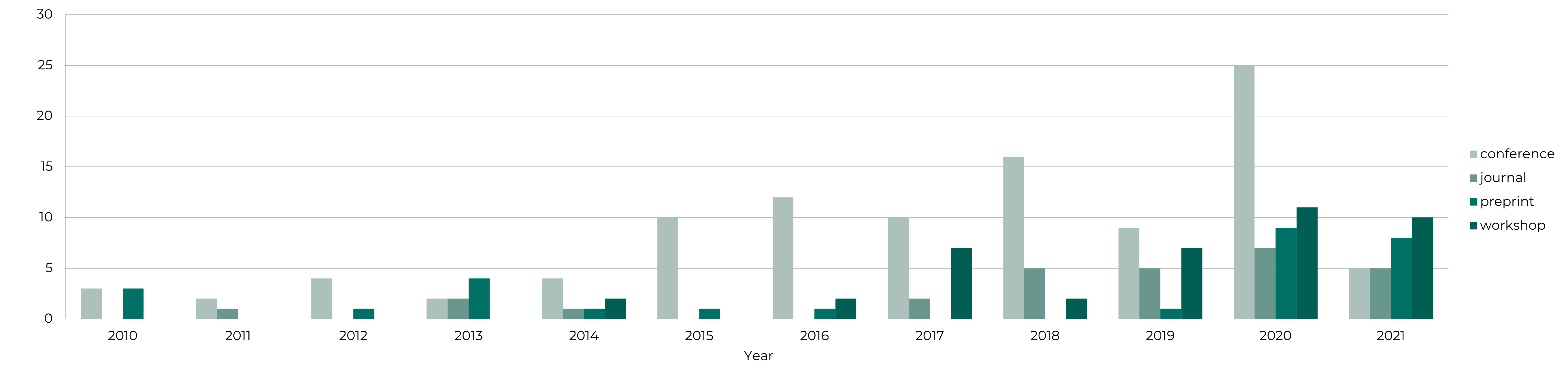}
\caption{The count of publications across conferences, journals, preprints and workshops.}
\label{fig:year_venue}
\centering
\end{figure*}

Figure \ref{fig:year_venue} depicts the evolution of Arabic NLP in the light of publicly available data resources from various venues. The graph demonstrates a general growth in the number of published resources, with a particular increase in even years. This can be attributed to the large number of datasets published at the bi-annual LREC conference.  We also anticipate a  significant increase, particularly in 2020, with the emergence of pretrained language models language models namely AraBERT \cite{antoun2020arabert}, Multi-dialect BERT \cite{talafha2020multi} and Araelectra \cite{antoun2020araelectra}. We can also observe that most of the datasets are published in conferences and workshops.


\subsection{Data Accessibility }
Data accessibility is an important aspect of fostering open research. Making the data source available extends its lifespan and allows it to be utilized in the way the dataset authors intended. In our initial data sources collection, we observed that more than half of the 93 of the discarded datasets had no online presence, nor an explicit means of accessing any version of the data. Based on the examination of the data sources, there is  general trend of making the data available through open source repositories such as GitHub, GitLab and Mendeley Data.
In the last three years, more than 80 \% of the data can be accessed freely on different data hosters. This trend is very promising as it shows an increased interest in making the datasets available online. In the recent years, there is a small portion of datasets that needs authentication to access either through email or registration forms. We also observed some papers that suggested contacting the corresponding author to access the data privately. We didn't include such papers on our \textit{final list}. 

\subsection{Data Providers and Licensing} 
Data providers are important to collect, annotate, distribute, and perhaps host the datasets. Another responsibility  of the data provider is to select the appropriate licence for the datasets. In fact, having a proper licence is a key component of any dataset for both data providers and researchers.  In terms of data providers, our findings show that the majority of the data sources we collected were created in virtue of collaboration of multiple institutions.  Institutions such as QCRI, Qatar university, NYU Abu Dhabi, and Nile University are the top four providers of Arabic NLP datasets. While datasets from these prominent providers are typically  accompanied by clear licensing. Unfortunately, about 50 \% of the datasets lack licences. Among those with explicit licence, there is a wide range of used licences. Some examples include several variations of Common Creative licences, Apache, MIT, GPL and BSD.

\subsection{Dialects Diversity}
\begin{figure}[htp]
\includegraphics[width=0.5\textwidth]{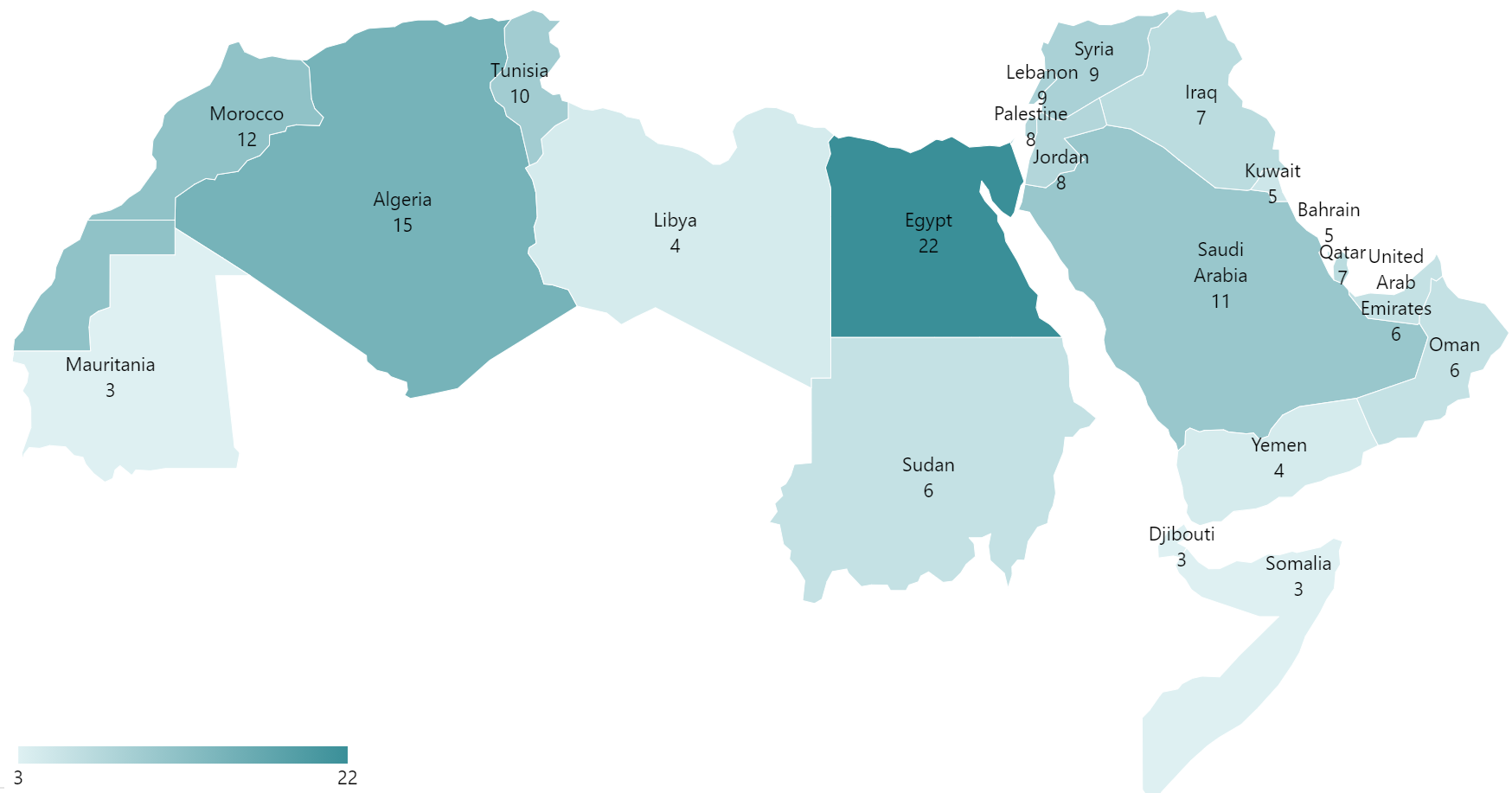}
\caption{Dialects representation across datasets.}
\label{fig:dialects_diversity}
\centering
\end{figure}

As we can see from Figure \ref{fig:dialects_diversity}, there were more than 20 entries out of the 200 with annotations of the dialects. These datasets are primarily intended for dialect identification tasks. The scope of the datasets we collected across the Middle East and North Africa is depicted in Figure. The Egyptian dialect is the most prevalent, followed by Algerian, Moroccan, and  Saudi dialects.  Somali, Djibouti, and Mauritanian dialects are underrepresented in the surveyed datasets, with only three resources for each.


\subsection{Tasks Coverage}
\begin{figure*}[htp!]
\includegraphics[width=1\textwidth]{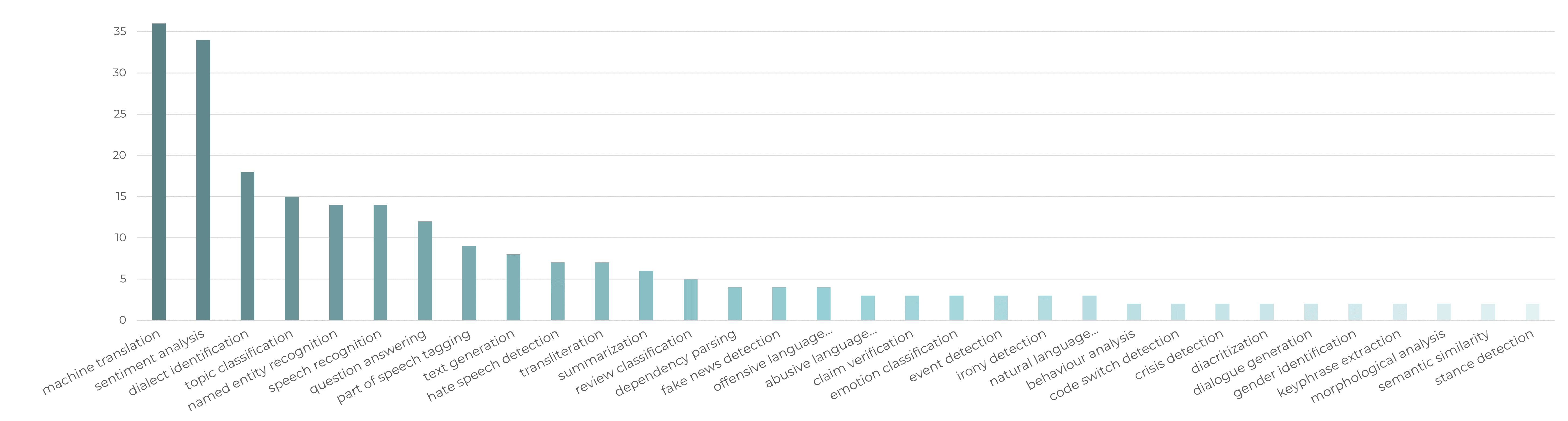}
\caption{Tasks' histogram. We only show the tasks that appeared more than once in papers}
\label{fig:tasks_histogram}
\centering
\end{figure*}

Figure \ref{fig:tasks_histogram} illustrates the distribution of tasks that appeared in more than one dataset.  The graph shows that machine translation and sentiment analysis are the most popular tasks within Arabic NLP community. Machine translation has received an increasing attention in  the literature across many languages, particularly in multilingual datasets, which explains the high frequency of publications in that area. Sentiment analysis, on the other hand, is heavily researched for a variety of reasons. Partly because the datasets for this task are primarily derived from social media sites with minimum effort, and partly because it serves as a suitable representation of everyday language that displays more sentiments. Other tasks that have presence within the community include dialect identification, topic classification, named entity recognition and speech recognition. There are also several low resource tasks that appeared at most once but are not displayed in the figure, such as poetry classification, word disambiguation, grammar checking, to name a few. Each of these sources only contains a single dataset addressing a distinct task. In contrast, datasets like KALIMAT \cite{el2013kalimat}, contains annotations for multiple tasks, or evaluation suites such as ALUE which is an aggregation of multiple datasets\cite{seelawi2021alue}.


%% file: 51_issue_and_recommendation.tex
\section{Arabic NLP Datasets Challenges and Recommendations}
\label{rq3}
This section highlights some issues of Arabic NLP data related to their legitimization, the haphazard collection, annotation, and the documentation practices. 

\paragraph{Data Availability.}

It is encouraging that our review identified 200 datasets that were publicly available, yet discoverability seems, by all accounts, to be an issue. While a few datasets are well recognized in the field, many are not, which might potentially lead to missed research opportunities and might result in bias because of an overuse of a few potentially non-representative datasets.  Further considerations in this regard, arising from our survey, include the sustainability (persistence) of the dataset URLs. As there is not a dedicated platform to host Arabic NLP datasets, some datasets' links appear to be inaccessible due to URLs invalidity (\textit{orphan datasets}). 
We identified one obvious/clear cause of orphaned datasets, which is the termination of academic affiliation and so the broken dataset link when the dataset is published as part of the researcher's academic webpage. One possible solution to address this issue is to host the datasets on public repositories like GitHub, Gitlab, Mendeley Data, SourceForge, to name a few.




\paragraph{Data Documentation.}
Data documentation refers to the process that describes the collected data and aims at facilitating cataloguing and discoverability of the data. One key form of data documentation is metadata, which are characteristics describing the data. For a dataset to be truly reusable, adequate documentation is important to offer the necessary insights into the potential usage of the dataset. For researchers, providing such insights saves time and resources, and it suggests a reliable dataset for reuse \cite{perrier2020views}.

In this work, we analyse the documentation of Arabic NLP datasets in relations to the  proposed metadata in Section \ref{methodology}. When we examine these datasets, we recognize that a few of them are accompanied by documentation. The majority appears to report inadequate metadata that is insufficient to make a decision on the dataset reuse. More precisely, there appears to be a pattern in which some researchers are satisfied with only publishing the direct URL link to download the dataset, or accompanying the downloadable dataset with a \texttt{README} file stating the size of the dataset, and a reference to the publishing paper on the dataset host page. Within the quality metadata, we observed very few instances reporting the data collection style. Another consideration includes the absence of clarity around the terms of access and use from some dataset providers. In most cases, we noticed that datasets are not accompanied by sufficient information regarding their provenance, and hence it is not possible for researchers to know if there is an appropriate ethical and governance framework underpinning the provision of these datasets. As a result of this research, we conclude that governance information, such as licencing, is a crucial part of the documentation and that, if not specified, the dataset's potential reuse may be limited.
Regardless, as noted earlier, ease of access and good documentation is an important driver for researchers. 
Therefore, deploying a framework for documenting NLP data, such as those proposed by \citet{bender-friedman-2018-data} and  \citet{gebru2018datasheets} is considered as a good step towards promoting data sharing. 


 


\paragraph{Data Sharing.}
Data sharing is positively seen  in the NLP community, with even top conferences are recognizing researchers who have shown a desire to share datasets. This process is usually volunteered, unless it is enforced internally by institutions and corporates measures. Within the Arab NLP community, we observed a high intention to support the research by sharing datasets. While some datasets are poorly documented and hardly accessible, others are well-prepared with clear documentation. 

We identified some datasets that are never published, or they are inaccessible, even where there was an intention to make them available, as declared in the formal publications. 
In terms of openness, we also observed a pattern in which some providers require a form of registration prior to sharing the datasets. Regarding sharing dataset links, as it was detailed in the data documentation, the unsustainability of dataset link poses a challenge, and hence data repository such as Github and Gitlab are usually adopted as a platform for sustainable data sharing.



\paragraph{Evaluation.}
NLP models are usually evaluated by training on specific tasks. In the literature, the test split is provided as an approach to evaluate models after training on the training split. In our metadata collection process, we observed that more than 60 \% of the datasets do not have predefined test splits. To mitigate that, researchers replicate the experiments by evaluating the old models again on a chosen random split of the data. As a result, a dataset's results will be incomparable across different NLP models.

\paragraph{Data Collection or Curation.}
Having stated annotation protocols and clear justification behind inter-rater agreement increases the reliability of the data. In the surveyed datasets, we have the following observations about the collection style. First, some crawling driven datasets lack any consideration of ethics and legal frameworks imposed by the platform from which the data is scraped, and the country of the data subjects. It also imposes an ethical risk by stating personal identified information. Secondly, when using machine translation to drive an Arabic version of a non-Arabic dataset, we observed missing information such as the translation models, verification process by native speakers, the reported errors, to name a few. 
Given the current quality of machine translation models, this approach in creating Arabic datasets opens many questions about the quality of the dataset and its potential usage.
While this approach can be used to help create datasets for some tasks, it is a risky approach if used to drive benchmark datasets for tasks such as common sense. Thirdly, as it was highlighted in previous points, not having an indication of the ethical risk of using datasets is a weak point in the Arabic NLP datasets. While each Arabic-speaking country has its own legislations and data protection acts \cite{IntellectualPropertyLawsoftheArabCountries}, it is important to flag any potential risk of using the datasets for future usage. 



\paragraph{Ethical Concerns and Privacy.}
Social media data,  such as Twitter data,  composed the greatest proportion of Arabic NLP datasets, particularly for dialect representations. Typically, such data is associated with the risk of exploiting personal information. In fact, this raises the concerns of considering data subject's right and the ethics behind using such data. Likewise, datasets acquired from publications and human-produced literature pose concerns regarding the incorporation of copyright consideration in the derived Arabic NLP datasets. In either type of the datasets, we found no explicit risk indications at any point of the NLP pipeline: collection, modelling, evaluation, or deployment. In this context, we encourage data providers to state information about the ethical risks associated with their released datasets, as well as the appropriate approach to mitigate them, in order to enrich the Arabic NLP landscape.

%% file: 6_limitations_and_future_work.tex
\section{Limitations and Future Work}
\label{ref:section-limitations}
We recognize some limitations to our study. Firstly, given the nature of our search strategy, only datasets that are probably indexed with metadata, and whose publication contains one of our search key-phrases are likely to have been retrieved. Secondly, there exists some additional data resources available that are either with open access or regulated access (e.g LDC), but they were not explored in this study since they do not conform to our inclusion criteria. As a future work, we plan to keep the catalogue updated by adding new datasets and also support community-based contributions where authors can submit the metadata of their datasets to our online catalogue.  

%% file: 7_conclusion.tex
\section{Conclusion}
In this research, we created an online catalogue of 200 Arabic NLP datasets with metadata annotations. We analyzed our findings, discovered some issues and suggested some resolutions. Mainly, we recognise that the NLP field is rapidly evolving, and that both Arabic NLP researchers and practitioners recognise the value of incorporating Arabic into language technologies, particularly beyond Modern Standard Arabic. As a result, while this research provides a comprehensive analysis, it is only a snapshot in time and extra efforts are required to drive the field more in that direction. 

%% file: 8_acknowledgements.tex
\section*{Acknowledgements}
This research was conducted under the BigScience \footnote{\url{https://bigscience.huggingface.co/}} initiative for open research, a one-year-long research initiative targeting the study of
large models and datasets. It was conducted as part of the data sourcing group for collecting datasets for different languages. We would like to thank the members of the data sourcing group for the insightful discussions. 